\documentclass{eeict}
\inputencoding{cp1250}
\usepackage{caption3}
\usepackage{float}
\usepackage{hyperref}

\title{Feature space reduction as a data preprocessing for the anomaly detection}
\author{Simon Bilik}
\programme{Doctoral Degree Programme (2), FEEC BUT}
\emails{bilik@feec.vutbr.cz}

\supervisor{Karel Horak}
\emailv{horak@feec.vutbr.cz}

\abstract{In this paper, we present two pipelines in order to reduce the feature space for anomaly detection using the One Class SVM. As a first stage of both pipelines, we compare the performance of three convolutional autoencoders. We use the PCA method together with t-SNE as the first pipeline and the reconstruction errors based method as the second. Both methods have potential for the anomaly detection, but the reconstruction error metrics prove to be more robust for this task. We show that the convolutional autoencoder architecture doesn't have a significant effect for this task and we prove the potential of our approach on the real world dataset.}

\begin{document}

\maketitle

\selectlanguage{english}
\section{Introduction}

Anomaly detection is a state-of-the-art task focused on the outlying samples recognition. This task is challenging especially because of the large possible variety of the anomalous samples, which are usually not present in the required amount or even not at all in the training data. The other problem, especially with the image data, is the high dimensionality of those samples, where the most of those features are not crucial for the anomaly detection itself. All of this makes the anomaly detectors difficult to use in a proper manner.

For this reason, we investigate several methods, such as convolutional autoencoders (CAE), principal component analysis (PCA) with the t-distributed stochastic neighbor embedding (t-SNE) and the reconstruction metrics in order to reduce the input data dimensionality. To the best of our knowledge, we are the first to examine the feature space created from the error metrics. We suggest the use of the One Class Support Vector Machine (OC-SVM) to automatically process the feature space.

\section{Related Research}

The papers \cite{Scholkopf} and \cite{Ratsch} define the backgrounds of the One Class SVM. It defines this problem as the SVM model applied on the unlabeled data and it describes the algorithm and its optimization in detail. The implementation of this method can be found besides others in the \textit{Novelty and Outlier detection} scikit-learn Python package. This method is quite efficient, but it often fails on the high dimensional data, as it is described in \cite{Erfani}. Despite this factor, it is usually used only as a classification layer for a reduced feature space, rather than for straight use with image data.

The article \cite{Masci} describes the CAE architecture principles. The authors describe general applications of the autoencoders, as image denoising, or feature space reduction. They also compare CAEs with fully connected autoencoders and describe the advantages of CAEs. In this article, authors also provide a mathematical background of the CAE and show the importance of the max-pooling layer for the learning consistency. A possibility of initializing the classification CNN with the same topology CAE is explained including its benefits.

The following paper \cite{Sakurada} compares performance of the fully connected AEs, denoising AEs and the Gaussian kernel PCA on the artificial multidimensional data and satellite's telemetry data. The authors prove that the AEs architectures can learn to represent normal data and that the reconstruction error could be used as an anomaly score even in the multidimensional data. The AEs based methods clearly outperform the PCA in this research.

The work \cite{Bergman} from the company MVTec presents a CAE based defect segmentation method. The authors compare several reconstruction loss metrics and structure similarity (SSIM) as L2 metric together with variational AE and feature matching AE architectures. The SSIM metric is concerned to bring the best results. The main advantage of this method is that not only the single pixel values are compared, but also the luminance, contrast and other parameters are reflected. The authors also provide an overview of the state-of-the-art CAE based anomaly detection methods and a description of the metrics mentioned above.

The article \cite{Vincent} shows a new training principle in order to extract robust features with the denoising AEs. The main idea of the presented approach is that a noise-damaged sample is processed by the AE as an input and the original sample is considered as a label. For the input corruption, the authors set a fixed number of randomly selected input elements to zero, but they suggest a use of other corrupting noise. This approach should help the AE to select stronger features and to better generalize the inputs.

\section{Materials and methods}

\subsection{Dataset description}

For this experiment, we used the \textit{Cookie Dataset}. It is our internal dataset designed for the anomaly detection task, which captures \textit{Tarallini} biscuits. It contains 1225 samples in four classes with the following structure:

 \begin{itemize}
    \item No defect (474 captions)
    \item Defect: not complete (465 captions)
    \item Defect: strange object (158 captions)
    \item Defect: color defect (128 captions)
 \end{itemize}
 
\begin{figure}[H]
\begin{center}
  \includegraphics[width=12cm,keepaspectratio]{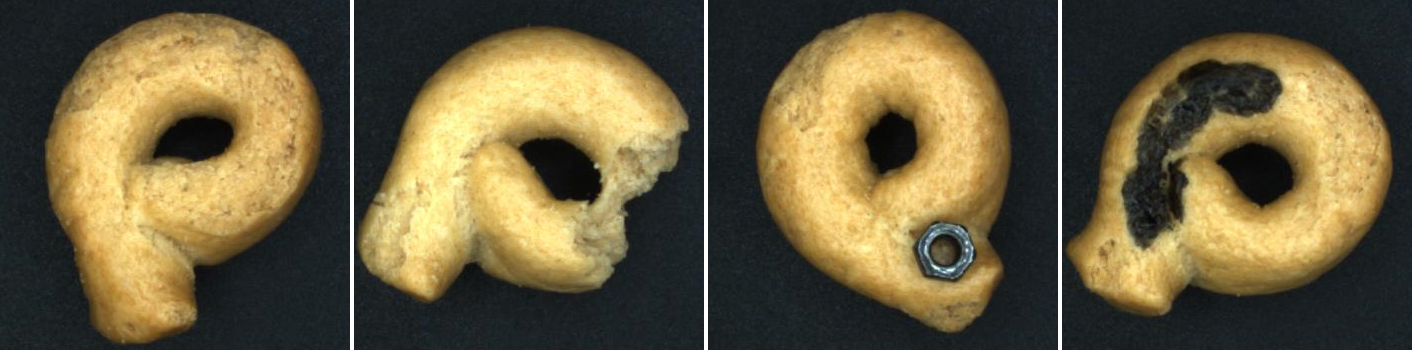}
  \caption{\textit{Cookie Dataset} class samples}
  \label{fig:dataset}
\end{center}
\end{figure}

All classes can be seen in fig. \ref{fig:dataset}. In order to augmentate the dataset we rotated each sample by 90\textdegree\, for three times, so the augmented dataset contains 4900 captures in total. Augmented images were cropped to the size of their bounding boxes, the source samples remained uncropped in the original dataset.

From this data set we created a subset for the anomaly detection, where we consider only the samples within specification (OK class) and randomly selected faulty samples from all faulty classes (NOK class). Our training dataset contains 1000 OK samples and no NOK samples as the training set, 2000 OK samples and no NOK samples as the validation set and with 200 OK samples and 200 NOK samples as the test set. All images were cropped to their bounding boxes. The sample ratio of the NOK classes in the test set was set as 0,4 : 0,3 : 0,3 (not complete : strange object : color defect).

\subsection{Experiment description}

We use CAE as a first stage of our pipeline. For this purpose, we compare the feature extraction quality of a basic CAE implementation (BAE1) with six convolutional layers\footnote{\url{https://blog.keras.io/building-autoencoders-in-keras.html}}, our own simplified version of the implementation mentioned above with four convolutional layers (BAE2) and the architecture suggested by MVTec \cite{Bergman} with sixteen convolutional layers.\footnote{\url{https://github.com/cheapthrillandwine/Improving\_Unsupervised\_Defect\_Segmentation}}

As the dimensionality of the input samples is still too high for the anomaly detection task, we compare two methods for the feature space dimensionality reduction. As the first method, we reduce the number of input features to 50 with the PCA algorithm and then to 2 with the t-SNE method as suggested in \cite{scikit_TSNE}. The second method we used was the reconstruction error metrics based feature space. For this purpose, we chose to use the L2 metric together with the SSIM suggested in \cite{Bergman} in order to create a two dimensional feature space.

For the PCA, t-SNE and OC-SVM methods, we used the implementations available from the \textit{scikit-learn} Python library \cite{SciKit}. All computations were performed on the Nvidia GTX2080 graphic card. The size of the input images was 256x256 px and the batch size of the CAEs was set to five for training. Scripts used for the experiments were written in Python 3.6 and the Tensorflow 2.4.1 library was used to train the CAEs.

\subsection{Experiment results}

\begin{figure}[H]
\begin{center}
  \includegraphics[width=8cm,keepaspectratio]{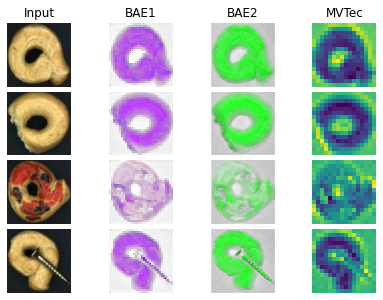}
  \caption{Visualisation of the CAE encoded inputs}
  \label{fig:CAE}
\end{center}
\end{figure}

The examined CAE architectures showed similar performances in the image reconstruction and a slightly different results in the image encoding, as can be seen in fig. \ref{fig:CAE}. The best input dimensionality reduction was achieved with the MVTec autoencoder. We tried to use both encoded and decoded samples as a direct input to the OC-SVM classifier, but the similarity of the samples was still too high and this method did not bring us any good results.

\begin{figure}[H]
\begin{center}
  \includegraphics[width=15cm,keepaspectratio]{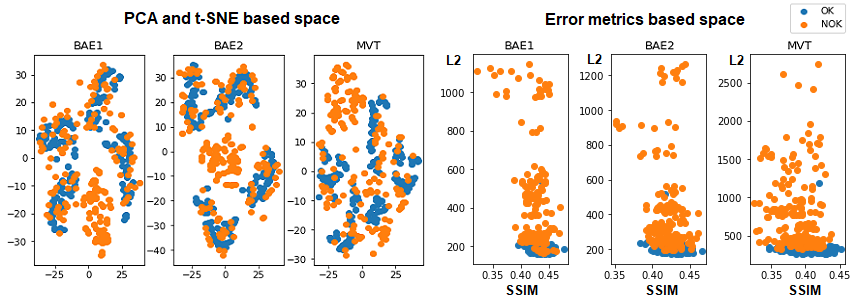}
  \caption{Comparison of the resulting feature spaces}
  \label{fig:TSNE}
\end{center}
\end{figure}

Results from the PCA and t-SNE based dimensionality reduction can be seen in the left part of fig. \ref{fig:TSNE}. We can see, that this method is able to separate one class of the NOK samples (probably the color defects), but the other NOK classes blend with the OK class due to their structural and shape similarity. The best results were obtained with the MVTec CAE, which has the lowest overlap of the OK and NOK class, although the OK class doesn't create a closed cluster. The disadvantage of this approach is that the t-SNE method has a stochastic behaviour and the feature space slightly differs after each run. We can imagine that this method could be used in case of more different OK and NOK classes.

The feature space based on the L2 and SSIM reconstruction error metrics performed better in the OK and NOK classes separation, although the OK class cluster still overlaps with a part of the NOK data. The results can be seen on the right part of figure \ref{fig:TSNE}. The overlapping NOK data are probably from the \textit{Not complete} class which is expected to show a lower reconstruction error than the other NOK class samples. Surprisingly and unlike in the paper \cite{Bergman}, the SSIM metric shows a low variance between the OK and NOK class, while the L2 metric seems to better distinguish those classes. It is probable, that the class separation might be increased by adding another reconstruction error metrics.


\section{Discussion}
Based on high dimensionality of the encoded samples, we assume that the convolutional autoencoders are not suitable for the stand-alone use, but they can be used as a part of the anomaly detection pipeline.

Feature space reduction using the encoded samples with PCA and t-SNE methods didn't perform as expected on the given dataset, when only one NOK class created a separated cluster. On the other hand, the results show that this method can be used in the cases where the NOK class differs more significantly from the OK class (eg. color defects rather then the shape, or small strange objects). The improvement of this approach will be an object of the following research.

The best results in the anomalous samples separation in the feature space were achieved using the L2 and SSIM reconstruction error metric, although the metric selection should be further studied and the classes separation is still far from ideal. We believe that better results might be obtained by increasing the feature space dimensionality with other error metrics, or by comparison of the reconstructed sample with the suitably selected and representative samples from the OK class. This approach might also lead to better results of the SSIM metric.

In all three examined cases we tried to use the OC-SVM classifier to determine the boundaries between the OK and NOK class. This method did not perform as well as expected - we plan to focus on tuning and comparison with other unsupervised classifiers in our future experiments. We also suggest the normalization of the feature space and experimentation with another error metrics because SSIM in particular did not fulfil our expectations. 

\section{Conclusion}

In this paper, we compare several approaches how to reduce the feature space for anomaly detection. Because the performance of all studied CAE architectures is similar - based on the lowest encoded space dimensionality, we would recommend the MVTec for the further use. We prove that the error metrics can be used to create feature space for the anomaly detection task and that this space is more suitable for this task than the feature space based on PCA and t-SNE. To the best of our knowledge, we are the first to use the error metric in this manner and not only as an anomaly score.

We also plan to test this approach on another dataset for anomaly detection, because the \textit{Cookie Dataset} shows low variance between the classes and especially the structural defects are a difficult task to distinguish.

%
\section*{Acknowledgement}
The completion of this paper was made possible by the grant No. FEKT-S-20-6205 -"Research in Automation, Cybernetics and Artificial Intelligence within Industry 4.0" financially supported by the Internal science fund of Brno University of Technology.

%

\end{document}